\documentclass[a4paper, 10pt, conference]{ieeeconf} 
\IEEEoverridecommandlockouts                              % This command is only needed if 
\usepackage[utf8x]{inputenc}

\usepackage{float}
\usepackage[caption = true]{subfig}
\usepackage{graphicx}
\graphicspath{{figures/}} %Setting the graphicspath   
\usepackage{footnote}
\makesavenoteenv{table}
\usepackage{algorithm}  
\usepackage{algorithmicx}
\usepackage{algpseudocode}
\usepackage{amsmath}
\usepackage{amssymb}%
\usepackage{booktabs}
\floatname{algorithm}{Algorithm}

\title{\LARGE \bf
	An Efficient L-Shape Fitting Method for Vehicle Pose Detection with 2D LiDAR 
}

\author{Sanqing Qu$^{1}$, Guang Chen$^{1,2,^{*}}$, Canbo Ye$^{1}$, Fan Lu$^{1}$, Fa Wang$^{1}$, Zhongcong Xu$^{1}$, Yixin Ge$^{1}$	
	\thanks{${1}$ College of Automotive Engineering, Tongji University}
	\thanks{${2}$ Chair of Robotics, Artificial Intelligence and Real-time Systems TUM Department of Informatics Technical University of Munich}
	\thanks{${*}$ Corresponding author E-mail address: {\tt\small guang@in.tum.de}}
}

\begin{document}

	\maketitle
	\thispagestyle{empty}
	\pagestyle{empty}
	
\begin{abstract}

Detecting vehicles with strong robustness and high efficiency has become one of the key capabilities of fully autonomous driving cars. This topic has already been widely studied by GPU-accelerated deep learning approaches using image sensors and 3D LiDAR, however, few studies seek to address it with a horizontally mounted 2D laser scanner. 2D laser scanner is equipped on almost every autonomous vehicle for its superiorities in the field of view, lighting invariance, high accuracy and relatively low price. In this paper, we propose a highly efficient search-based L-Shape fitting algorithm for detecting positions and orientations of vehicles with a 2D laser scanner. Differing from the approach to formulating L-Shape fitting as a complex optimization problem, our method decomposes the L-Shape fitting  into two steps: L-Shape vertexes searching and L-Shape corner localization. Our approach is computationally efficient due to its minimized complexity. In on-road experiments, our approach is capable of adapting to various circumstances with high efficiency and robustness.

\end{abstract}
\section{Introduction}
Nowadays, autonomous driving has become one of the most attractive and cutting edge topics. Although there is still a lot of work to do before the arrival of fully autonomous driving, semi-autonomous driving is already accomplished and will be widely introduced in the near future. For semi-autonomous driving vehicles, it is necessary to have the ability of avoiding obstacles to ensure the driving safety. The surrounding vehicles' locations and orientations detection is very important in collision avoidance.

Light Detection And Ranging (LiDAR) has been widely used for detecting surrounding objects such as bicycles, vehicles and pedestrians, due to its large field of view, lighting invariance, high data accuracy and relatively low price. A common approach to processing LiDAR data is to segment the data into different clusters of points, from which meaningful features like line segments, rectangles, and circles can be extracted~\cite{refer1}. These features are then associated with a static map or tracked targets and used to update the target state through tracking methods such as Multiple Hypotheses Tracking (MHT)~\cite{MHT_1,MHT_2} or its advanced version which integrates a Rao-Blackwellized Particle Filter (MHT-RBPF)~\cite{MHT_RBPF_1,MHT_RBPF_2}.

Another solution is similar to the approach widely used in computer vision by extracting hand-crafted features and training classifiers. Image-based object detection is very popular in current autonomous driving research, such as road obstacles detection~\cite{ppp18}, mobility aids~\cite{ppp34} and vehicle detection~\cite{ppp28}. However, the sparse point data from a 2D LiDAR are usually insufficient for reliable object identification using this kind of method within a single scan. Although several solutions are proposed, such as relying on sensor fusion~\cite{ppp30}, multilayered sensor combinations~\cite{ppp20},~\cite{ppp29}, or temporal integration from tracking, they often come with higher computational cost and complexity.

\begin{figure}[t]
	\centering
	\subfloat{\includegraphics[width= 3.3in]{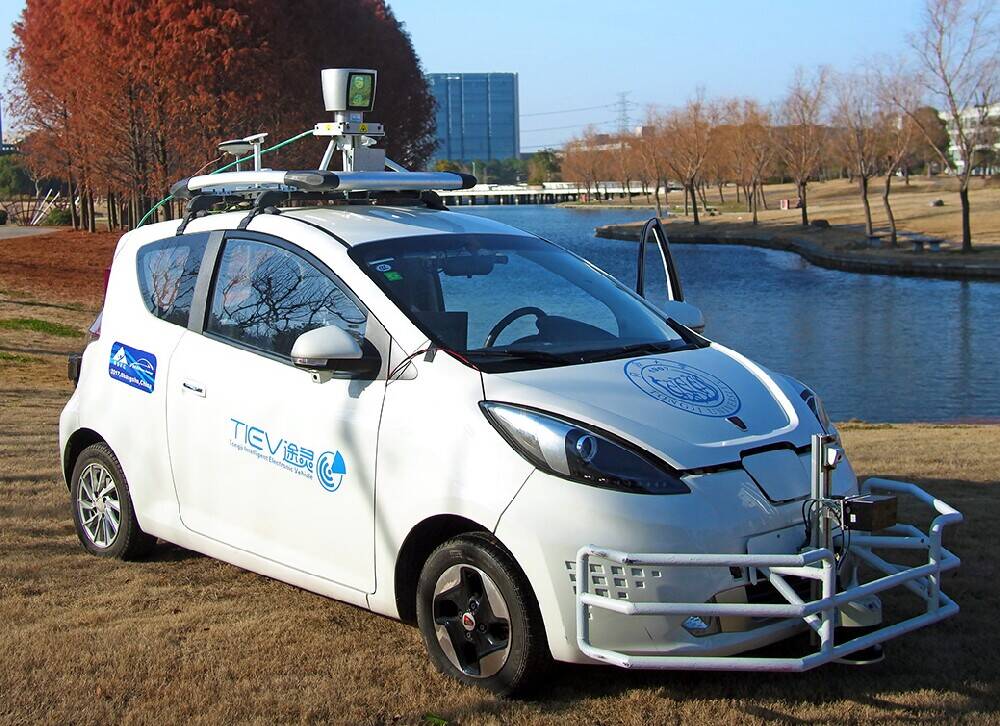}}
	\caption{Tongji autonomous vehicle research platform “TIEV”}
	\label{figs:tiev}
\end{figure}

In this paper, we propose a highly efficient search-based L-Shape fitting algorithm for detecting the vehicle's position and orientation. L-Shape fitting is often treated as a complex optimization problem. However, our approach addresses this problem by decomposing it into two steps: L-Shape vertexes searching and L-Shape corner localization. It is extremely important to ensure the real time performance of vehicle detection and to save time for highly computational tasks such as high-level path planning and decision making tasks. Our method is demonstrated to be effective and efficient through experiments with a production-grade 2D laser scanner.

The remainder of this paper is organized as follows. In section II an overview of related research work is described. The searched based L-Shaped fitting method is presented in section III. In section IV, we provide the experimental results to evaluate the L-Shape fitting approach. Section V presents our conclusion from the experimental results.

\section{Related Work}
In the past decade, the well-known DARPA grand challenge has proved the realizability and demonstrated the technical frameworks for autonomous driving. Supported by NSFC (the National Natural Science Foundation of China), China's event named  the Intelligent Vehicle Future Challenge (IVFC), which is similar to DARPA urban challenge, started from 2009. During the last eight years, over thirty universities and many companies have participated in this annual challenge, which is now recognized as the most influential event of autonomous driving in China. As a latecomer in IVFC, Tongji Intelligent Electric Vehicle (TiEV) project funded by the Tongji University started in 2015 (see the Fig~\ref{figs:tiev}). Soon afterwards, TiEV took part in IVFC 2016, 2017 and managed to complete most of the tasks such as simulated traffic driving, going through tunnels and blockage avoiding without any human intervene. 

In the competition, the detection of curbs and tracking of vehicles were made possible using
the equipped sensors, such as cameras and laser scanners. However, from laser-based range sensing, we can only detect the parts of the object’s contour that faces towards the sensor. Since the contour of an object may not be fully observed by range sensors, these occlusions make the perception task of the autonomous vehicle even harder.

\par To address these difficulties, the vehicular shape model is widely used for detecting the position and orientation of the vehicle, which is often assumed to be a box, an L shape, or two perpendicular lines ~\cite{NUS_LShape,L_Shape_model_2,L_Shape_model_3}. Based on the vehicular L-Shape model, several fitting methods have been proposed. In~\cite{L_Shape_model_2}, a weighted least-squares method is used to get rid of outliers and fit an incomplete contour to a rectangle model. Because of the occlusion problem, both a right angle corner fitting and a line fitting are represented in~\cite{L_Shape_model_2}. In~\cite{NUS_LShape}, the information of the scanning sequence is exploited to segment the points efficiently into two disjoint sets, then two perpendicular lines corresponding to the two edges of the vehicle are fitted by each of the two segmentations of points respectively. More specifically, a pivot is detected based on the scanning sequence of all these 2-D range points, and then point of this pivot yield those two disjoint sets, i.e., the set of points scanned before the pivot and the set of points scanned after it. In~\cite{CMU_LShape}, the laser scanning sequential information is not utilized for L-Shape fitting. This method is based on the optimal fitting angle searching for the L-Shape, and in~\cite{CMU_LShape} three criteria were proposed to detect the best L-Shape fitting.

Also some other approaches were developed using volumetric data with 3D LiDARs, among which some choose sequential projections of point clouds~\cite{ppp7},~\cite{ppp8}, others choose to train up neural networks that can cope with unordered point cloud data with abstract feature learning, like in VoxelNet and PointNet. However, these approaches consume considerable computational resources and need a large-scale labeled data set for training, not to mention the sensors themselves are much more expensive than those for 2D ranging.

Compared with the methods above for vehicular shape fitting, we proposed a different approach to address the problem. There are four main contributions in this paper.
\begin{itemize}
	\item{We proposed an approach that innovatively decomposes the L-Shape fitting problem into two steps: L-Shape vertexes searching and L-Shape corner point locating. } 
	\item{The proposed approach is highly computationally efficient due to its minimized complexity, outperforms other methods and obtains state-of-the-art results.}
    \item{The proposed approach is robust enough and able to accommodate various situations.}
 \item{Our method does not depend on the laser scanning sequential information, which means data fusion can be easily achieved from multiple laser scanners.}
\end{itemize}

\iffalse
Firstly, our method also does not depend on the laser scanning sequence, which means that it's easy to achieve data fusion from multiple laser scanners. Secondly, our approach is computationally efficient thanks to its low-complexity. Thirdly, our method is robust, i.e., our method is capable of accommodating various situations.
\fi
\section{L-Shape Fitting for Laser Scanning Data}

\begin{figure*}
	\centering
	\subfloat{\includegraphics[width = 5in]{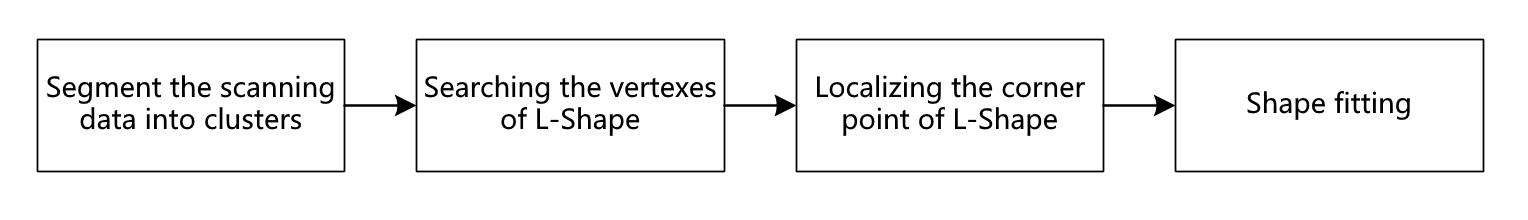}}
	\captionsetup{font={small}}
	\caption{The flow chart of this proposed fitting method.}
	\label{figs:flowchart}
\end{figure*}

Since the correspondence of the scanning data of the objects in the real world is usually complex, we first segment the data points into different clusters after getting the 
scanning data of the environmental objects using 2D LiDARs. These clusters typically correspond to bicycles, pedestrians, buildings, or vehicles and can be classified into separated categories. In this paper, we are only interested in L-Shape fitting for vehicles. Based on the assumption of an L-Shape vehicle model, for each segmented range data cluster, we first find the 2 vertexes (not including the corner point) of L-Shape and then localize the corner points based on a pre-specified criterion. After that, we obtain the optimized fitted rectangle following the 3 vertexes and contain all the points in this segmentation. Fig.~\ref{figs:flowchart} shows the flowchart of our approach.

\subsection{Segmentation}
The laser scan data needs to be segmented into different clusters before performing L-Shape fitting. There are several classical clustering algorithms for this segmentation work. For this work, we evaluate two classical clustering methods: mean-shift clustering (Mean-Shift)~\cite{clustering_meanshift} and density based spatial clustering of applications with noise (DBSCAN). The mean-shift algorithm considers the input as a probability density function and the objective of the algorithm is to find the modes of this function~\cite{clustering_meanshift}. These modes represent the centers of the discovered clusters. The input points are fed to the kernel density estimation and then the gradient ascent method is applied for the density estimate. The density estimation kernel uses two inputs: the total amount of points and the bandwidth or the size of the window~\cite{meanshift_And_dbscan}. The DBSCAN algorithm uses density based spatial clustering for applications with noise. For each point, the associated density is calculated by counting the number of points in a search area of specified radius, $\epsilon$, around the point. The points with density higher than the specified threshold value, MinPts, are classified as core points while the rest are classified as non-core points. 

\begin{figure}
	\centering
	\subfloat[DBSCAN ($\epsilon = 0.85, MinPts = 6$)]{\includegraphics[width = 3.2in]{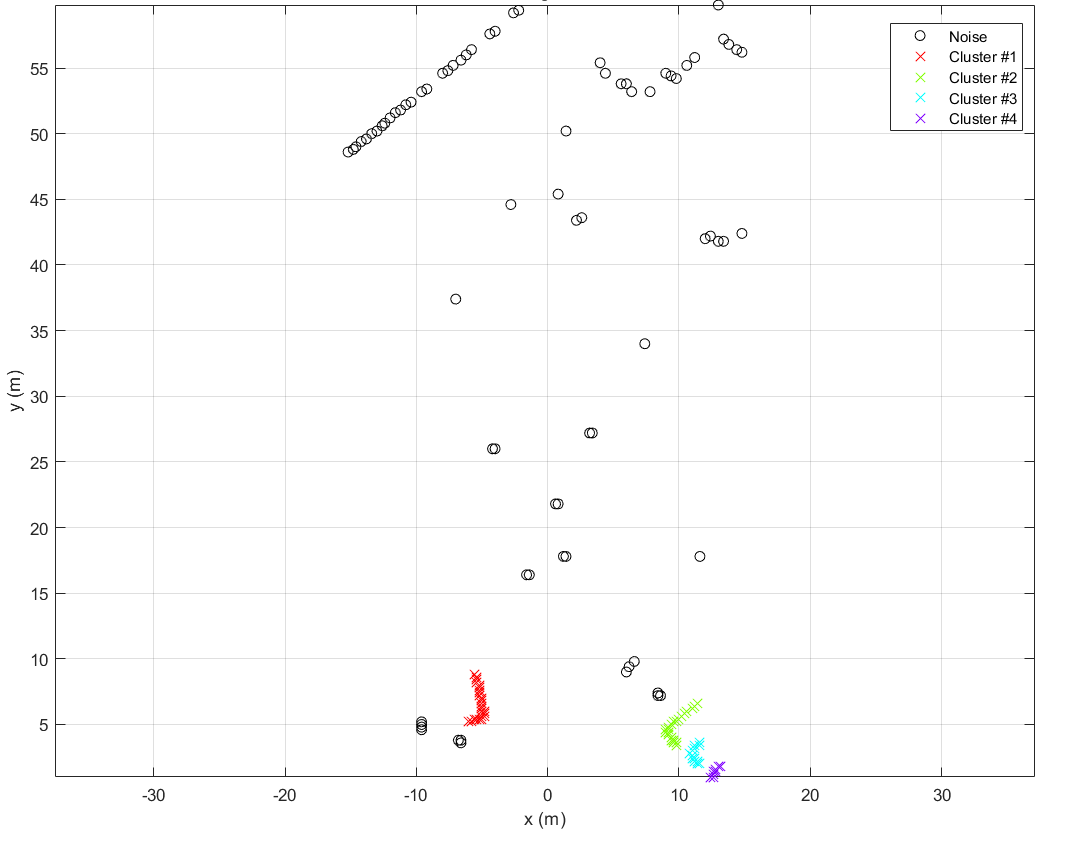}}
	\\
	\subfloat[Mean-Shift($Bandwidth = 6$)]{\includegraphics[width = 3.2in]{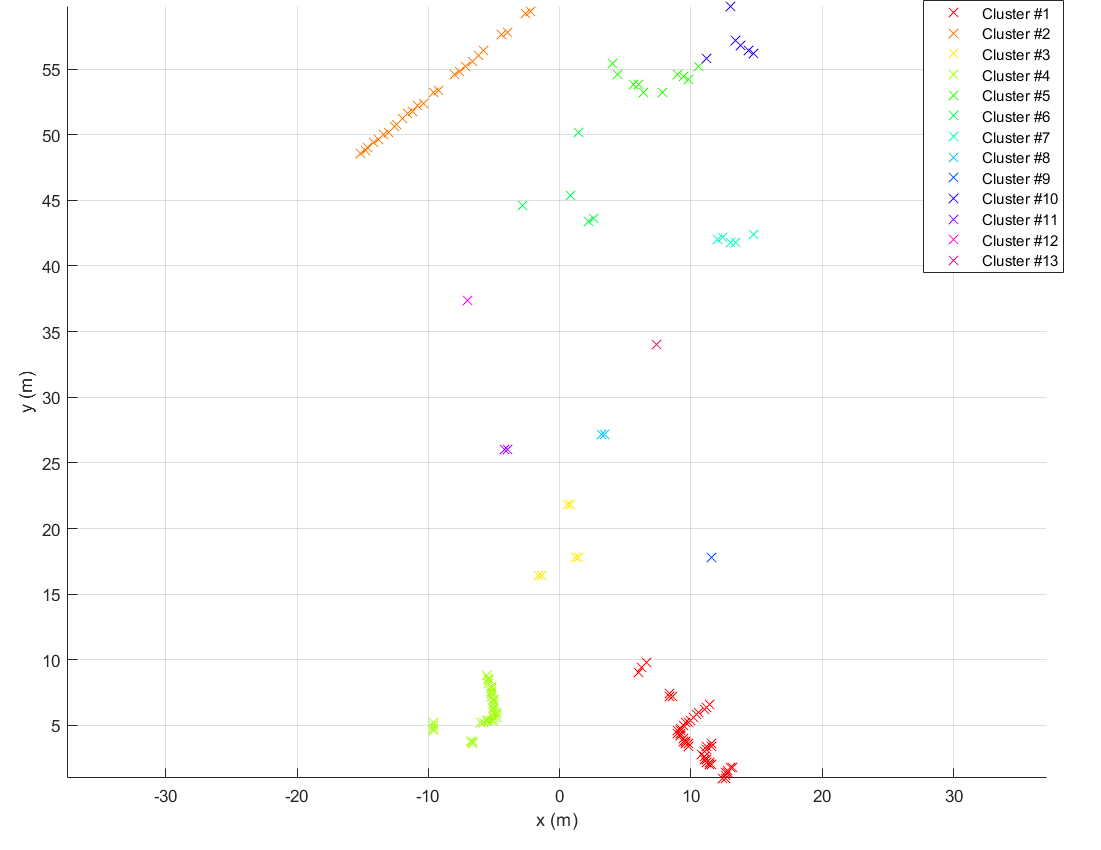}}
	\captionsetup{font={small}}
	\caption{Comparison of two classical clustering algorithms.}
	\label{figs:clustering_comparison}
\end{figure}

By comparing the segmentation results in Fig.~\ref{figs:clustering_comparison}, we can see
that both the DBSCAN algorithm and the mean-shift algorithm are able to find the clusters of arbitrary shapes. However, the mean-shift algorithm is not capable of ignoring the  influences of outliers. Furthermore, its iterative nature and density make the mean-shift algorithm slower than some alternative clustering algorithms. For these reasons, we used the DBSCAN algorithm to perform the segmentation due to its low-complexity, fast execution time and robust nature. It is worth mentioning that a graph-based index structure can be used to speed up the segmentation operation with the DBSCAN algorithm.%~\cite{DBSCAN_Speedup}

\subsection{L-Shape Fitting}
 Since the two perpendicular lines of L-Shape can be defined as $xcos\theta + ysin\theta = c_1$ and $-xsin\theta + ycos\theta = c_2$, a typical way to evaluate the fitting performance is least squares, which covers the following optimization problem: 
 \begin{equation}
 \left\{\begin{array}{l} 
 \min{\|\psi\| + \|\phi\|} \\[0.2cm]
 \mbox{subject to:}\\[0.1cm]
 \qquad \phi = A\cdot u_1\\
 \\
 \qquad \psi = B\cdot u_2\\
 \\
 \qquad  A = \begin{bmatrix}
 x_{P1} & y_{P1} & 1 \\
 x_{P2} & y_{P2} & 1 \\
 \vdots & \vdots & \vdots \\
 x_{Pp} & y_{Pp} & 1
 \end{bmatrix}
 u_1 = \begin{bmatrix}
 cos\theta\\
 sin\theta\\
 -c_1
 \end{bmatrix}\\
 \\
 \qquad  B = \begin{bmatrix}
 x_{Q1} & y_{Q1} & 1 \\
 x_{Q2} & y_{Q2} & 1 \\
 \vdots & \vdots & \vdots \\
 x_{Qq} & y_{Qq} & 1
 \end{bmatrix}
 u_2 = \begin{bmatrix}
 -sin\theta\\
 cos\theta\\
 -c_2
 \end{bmatrix}\\
 \\
 \qquad c_1,c_2 \in R \qquad	 0 \leqslant \theta \leqslant \frac{\pi}{2}\\
  \\
 \qquad P\cup Q = S, P\cap Q = \O
 \end{array}\right.
 \end{equation}
in which the optimization task is to find out two best partitions ($P,Q$) for the clustered preprocessed data $S$ and the optimal parameters for two orthogonal lines ($\theta, c_1, c_2$). The $\|\|$ means the $l_2$ norm, and $p,q$ are the scanning points' quantity for the partitions ($P,Q$).
\par Nevertheless, the above optimization problem turns out to be very difficult to solve due to the combinatorial complexities in partition since the order/sequence of points of the segmented cluster of is not accessible.
\par To address this computational problem, a basic idea is to implement RANSAC algorithm, since an L-Shape can be described with 3 key points. However, this original RANSAC algorithm also consumes plenty of time due to the considerable possibilities. To improve the algorithm's performance based on the 3 points theory mentioned above, we decompose the L-Shape fitting problem into two steps. L-Shape vertexes searching and L-Shape corner point localizing.
\subsubsection{Detecting Two Vertexes}
\par As the first procedure, we proposed an algorithm to obtain two vertexes of L-Shape from clustered scanning data. The algorithm is presented in the Alg.~\ref{Args:Vertexes searching}. The input of this algorithm is a specific cluster scanning points $S$ and the output are two target vertexes $V_A, V_B$ of L-Shape. It's worth mentioning that for improving the robustness of vertexes searching algorithm, the results are not actual several scanning points but the geometric center of specific points. 
\par
\begin{figure}[t]
	\centering
	\subfloat{\includegraphics[width= 3.3in]{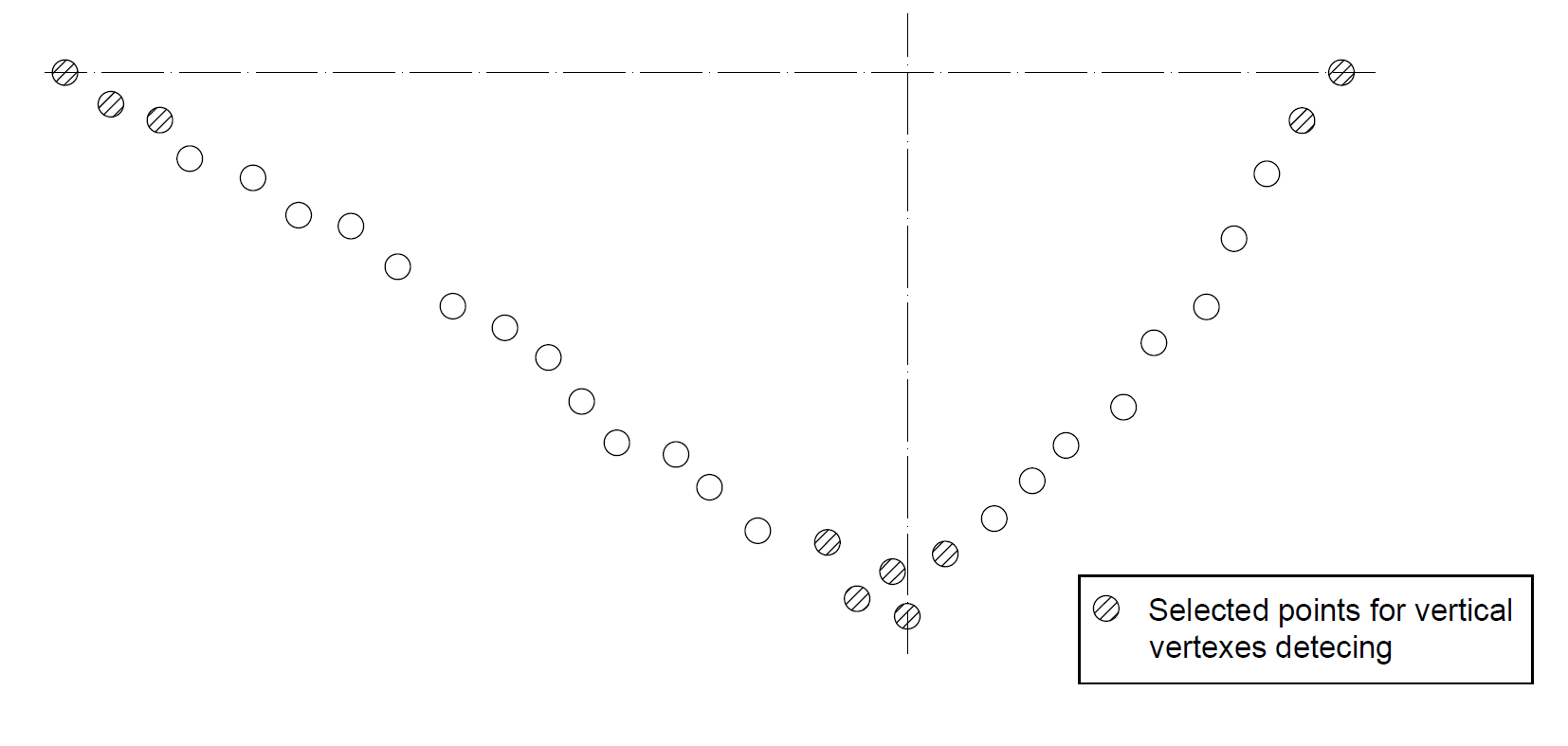}}
	\caption{special circumstance in vertex detecting}
	\label{figs:vertex_detect}
\end{figure}

 Firstly, we sort the points by their X and Y coordinates, and subsequently, we select several points from the front end and rear end of the sorted sequence and calculate the geometric center as candidate vertexes of L-Shape. After that, we based on a predefined standard to obtain two target vertexes.
In some scenes, the first or last several points may have a large variance in the horizontal or vertical direction (as in Fig.~\ref{figs:vertex_detect}). Under these circumstances, we can directly select the two calculated candidate vertexes of orthogonal direction as the L-Shape's two vertexes to reduce the computational cost.
%% This algorithm is better.
\begin{algorithm}[t]
	\caption{searching two vertexes of L-Shape}
	\label{Args:Vertexes searching}
	\begin{algorithmic}[1]
		\Require{ $n$ points in this cluster $S \in R_{n\times2}$}
		\Ensure{$V_A,V_B$, two vertexes of L-Shape}
		\State \mbox{sort points $S$ by abscissa as $X$}
		\State \mbox{sort points $S$ by ordinate as $Y$}
		\State get the four vertexes named $V_L,V_R,V_U,V_D$ of this points cluster using $X,Y$ 
		\State from $V_L,V_R,V_U,V_D$ select two almost superposed points $V_{s1},V_{s2}$, and mark the remaining vertexes as $V_{r1},V_{r2}$
		\State set the geometric center of $V_{s1},V_{s2}$ as $V_A$
		\If{$\angle V_A V_{r1} V_{r2} > \angle V_A V_{r2} V_{r1}$ }
			\State $V_B = V_{r2}$
		\Else
			\State $V_B = V_{r1}$
		\EndIf	
		\State \Return $V_A,V_B$
	\end{algorithmic}
\end{algorithm}
\subsubsection{Localizing Corner Point}
When the “Vertexes Searching”  procedure is completed, the second step for L-Shape fitting is to localize the corner points. Once this optimal corner point is obtained, the L-Shape feature for vehicle tracking is almost determined. 
A classical standard to evaluate fitting result has been presented at the beginning of “L-Shape Fitting”  section, minimizing the squared error. 
\par As the two vertexes have been determined, a basic idea is to traverse all the scanning points to localize the optimal corner point. Note that the optimal corner point can usually form an angle of approximately $90^\circ$ with the two given vertexes obtained from the Alg.~\ref{Args:Vertexes searching}. Therefore, a prejudgment procedure can be implemented for the scanning points to filter out some candidate corner points, before the localizing algorithm implemented to the candidate points.
\par The detailed algorithm is showed in Alg.~\ref{Args:corner_point_seeking}. The input of this algorithm are two vertexes and the corresponding points cluster, $S_{n\times2} \in R$. The output is the optimal corner point $P_{best}$ and the points' amount $N_{E_{1}},N_{E_{2}} $ of two disjunctions which were partitioned by the two vertexes and the optimal corner point. 
\begin{algorithm}[t]
	\caption{localizing optimal corner point of L-Shape}
	\label{Args:corner_point_seeking}
	\begin{algorithmic}[1]
		\Require{
			\mbox{$n$ points in this cluster $S=\{P_1,P_2,\cdots,P_n\}$}
			and two vertexes $V_A,V_B$ of the L-Shape}
		\Ensure{the optimal corner point $P_{best}$ and the points' amount $N_{E_{1}},N_{E_{2}} $ of the two disjunctions for cluster $S$.}
		\State init $Err_{min} = inf$
		\State init $N_{E_{1}},N_{E_{2}} =0$
		\For{$i=1,2,3,\cdots,n$} 
		\If{$\pi/2+\Delta\theta_0 \ge \angle{V_A P_i V_B}  \ge \pi/2-\Delta\theta_0$}
		\State \mbox{init $N_{temp1}, N_{temp2} = 0$}
		\State \mbox{init $Err = 0$}
		\For{$j=1,2,3,\cdots,n$}
		\If{$j\ne i$}
		\State $dis_1 = Dis(P_j,V_AP_i)$
		\State $dis_2 = Dis(P_j,P_i V_B)$
		\If{$dis_1 < dis_2$}
		\State $N_{temp1} ~ += 1$
		\State $Err ~ += dis_1$	
		\Else
		\State $N_{temp2}~ += 1$
		\State $Err ~ += dis_2$	
		\EndIf

		\If {$Err \le Err_{min}$}
		\State $Err_{min} = Err, P_{best} = P_j$
		\State $N_{E_{1}} = N_{temp1}, N_{E_{2}} =N_{temp2}$
		\EndIf
		\EndIf
		\EndFor 
		\EndIf
		\EndFor
		\State \Return $P_{best},N_{E_{1}},N_{E_{2}}$	
	\end{algorithmic}
\end{algorithm}

\subsection{Shape Fitting}
Since there is no ideal range data point, most angles formed by two vertexes obtained from Alg.~\ref{Args:Vertexes searching} and the optimal corner point acquired from Alg.~\ref{Args:corner_point_seeking} is actually not a real right angle. A logical idea is to select an edge which has more scanning points to determine the L-Shape's direction. As the Alg.~\ref{Args:corner_point_seeking} can return the two edge's points amount and the optimal corner point, with this information and the two vertexes obtained from Alg.~\ref{Args:Vertexes searching} the L-Shape's direction can be easily determined.

\par We use a rectangle oriented in that direction which contains all the scanning points to represent the L-Shape. Once this rectangle is obtained, the vehicle's pose can also be handily extracted. Since a rectangle is formed by four edges and every edge can be presented in the form of $ax+by+c = 0$, if these parameters are determined the Shape is acquired. The Alg.~\ref{Args:shape_fitting} shows steps about rectangle fitting in detail. The input of this algorithm are two vertexes $V_A, V_B$ obtained from Alg.~\ref{Args:Vertexes searching}, corner point $P$ and two partition points' amount $N_{E_{1}},N_{E_{2}}$ acquired from Alg.~\ref{Args:corner_point_seeking}. The output of this algorithm are the parameters for four edges of the target rectangle.
\begin{algorithm}[t]
	\caption{Shape Fitting}
	\label{Args:shape_fitting}
	\begin{algorithmic}[1]
		\Require{two vertexes $Ver_A, Ver_B$, corner point $P$,  two partition points' amount $N_{E_{1}},N_{E_{2}}$ and the 
		$n$ points in this cluster $S \in R_{n\times2}$ }
		\Ensure{ rectangle edges \{$a_ix + b_iy = c_i | i = 1,2,3,4$\}}
		\If{$N_{E_{1}} > N_{E_{2}}$}
			\State \mbox{$\theta_1 = atan2(V_{A_y} - P_y,V_{Ax} - P_x)$}
			\State \mbox{$\theta_2 = \theta_1 + \pi/2$}
		\Else
			\State \mbox{$\theta_2 = atan2(P_y - V_{B_y}, P_x - V_{B_x})$}
			\State \mbox{$\theta_1 = \theta_2 + \pi/2$}
		\EndIf
		\State $\vec{p_1} = (cos\theta_1, sin\theta_1)$
		\State $\vec{p_2} = (cos\theta_2, sin\theta_2)$
		\State $C_1 = S\cdot\vec{p_1}'$
		\State $C_2 = S\cdot\vec{p_2}'$ 
		\State $a_1 = cos\theta_1,\quad b_1 = sin\theta_1,\quad c_1 = min\{C_1\}$
		\State $a_2 = cos\theta_2,\quad b_2 = sin\theta_2,\quad c_2 = min\{C_2\}$
		\State $a_3 = cos\theta_1,\quad b_3 = sin\theta_1,\quad c_3 = max\{C_1\}$
		\State $a_4 = cos\theta_2,\quad b_4 = sin\theta_2,\quad c_4 = max\{C_2\}$
	\end{algorithmic}
\end{algorithm}
\section{Experimental Results}
In this section, we provide the experimental results to evaluate the correctness and efficiency of our algorithms. The experiments were tested on Tongji's autonomous vehicle research and test platform ``TIEV'' (in Fig.~\ref{figs:tiev}), and the 2D LiDAR is mounted on the front end of the test platform and about 15 cm above the ground with an elevation angle of about $1.5^\circ$. Under this circumstance, most of the vehicles in the measurement range are scanned as L-Shape. It is important to note that the scanning order/sequential information is not used for the experiments here.

\subsection{Rectangle Fitting}
\begin{figure}
	\centering
 	\subfloat{\includegraphics[width = 3.6in]{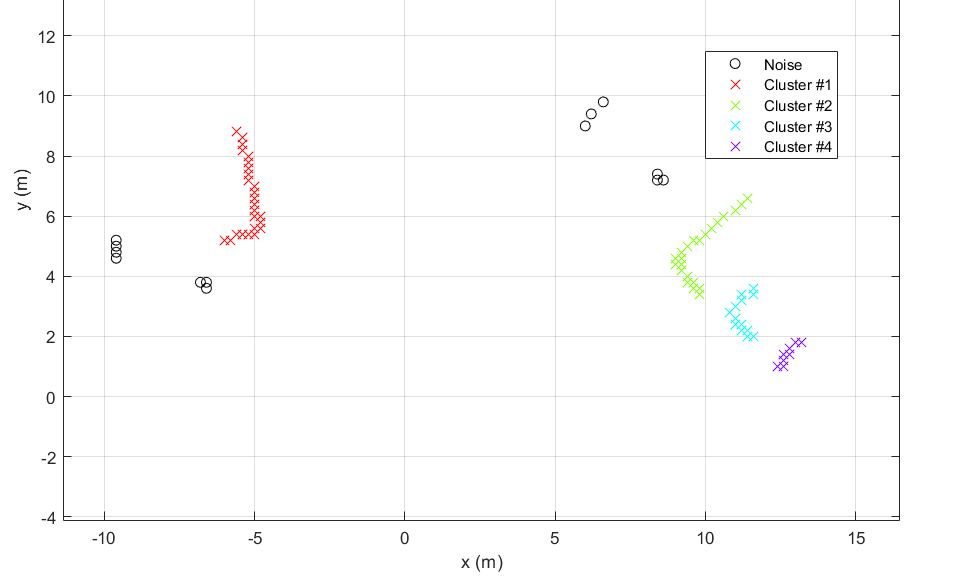}}
	\captionsetup{font={small}}
	\caption{The segmentation result for 1 frame laser scan data.(best viewed in color.)}
	\label{figs:DBSCAN_result}
\end{figure}

\begin{figure}
	\centering
	\subfloat[]{\includegraphics[width= 1.6in]{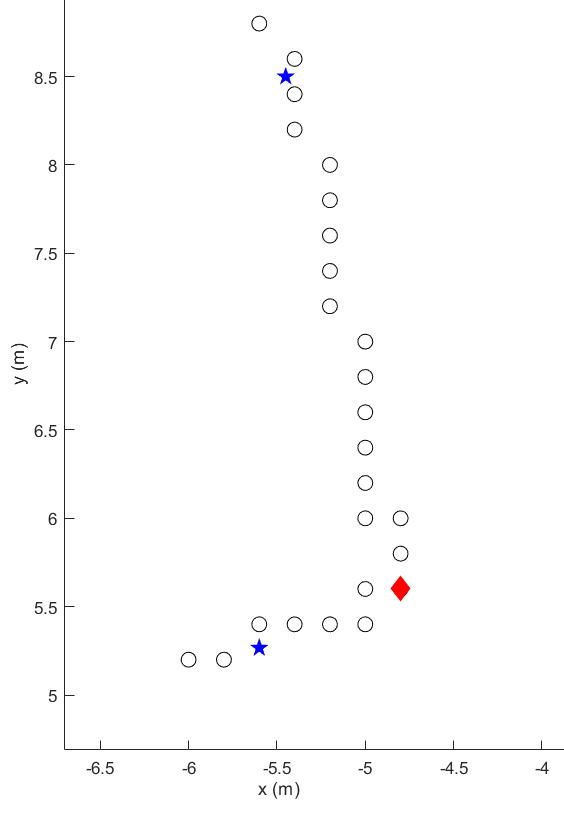}}
	\subfloat[]{\includegraphics[width= 1.6in]{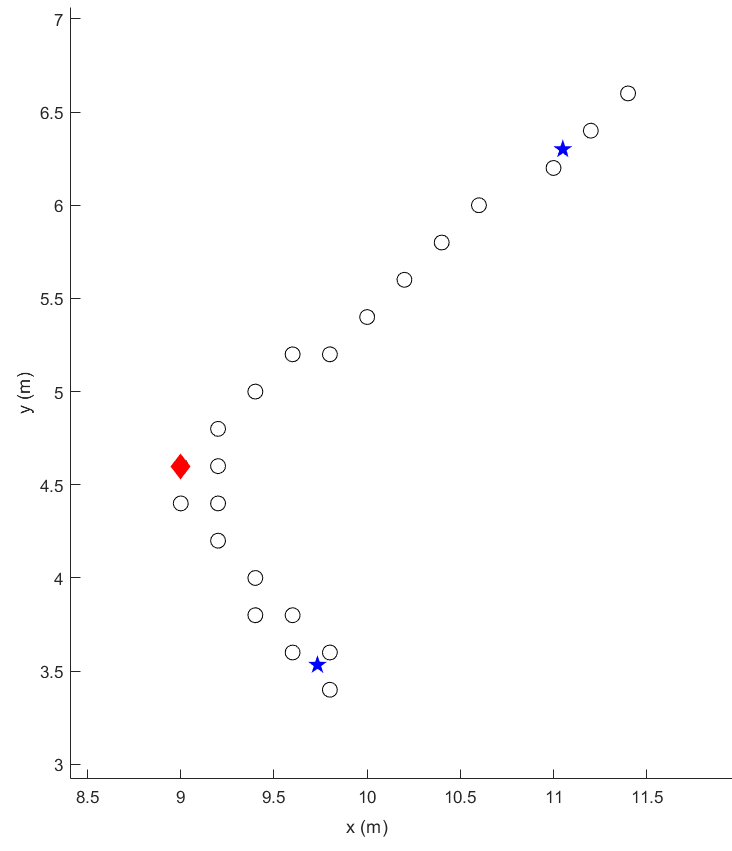}}
	\\
	\subfloat[]{\includegraphics[width= 1.6in]{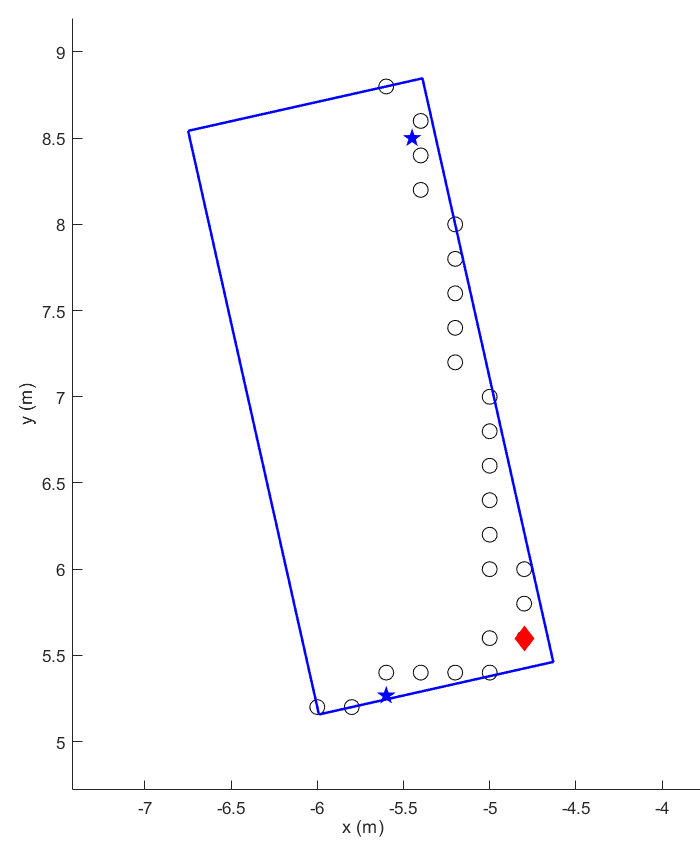}}
	\subfloat[]{\includegraphics[width= 1.6in]{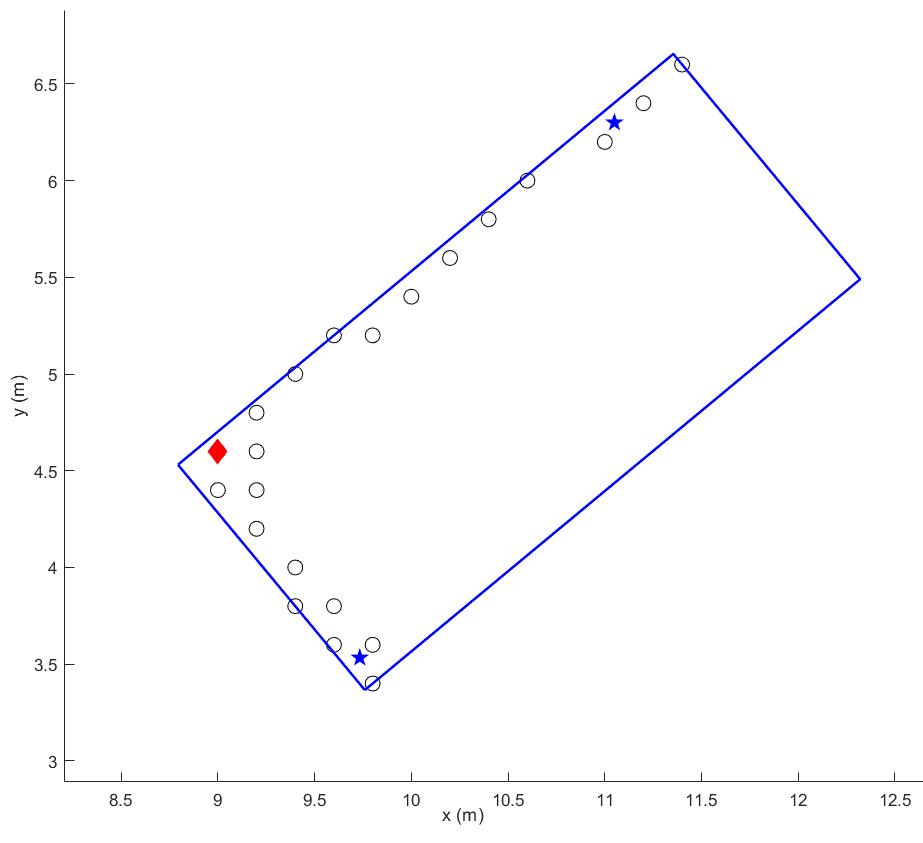}}
	\captionsetup{font={small}}
	\caption{\quad The key points and optimal L-Shape fitting results for two typical segmentations points clusters from the laser scanner. The blue stars in (a) and (b) represent the vertexes for L-Shape obtained from Alg.~\ref{Args:Vertexes searching} and red diamonds in (a) and (b) stand for the corner points for L-Shape acquired from Alg.~\ref{Args:corner_point_seeking}. Fig. (c) and Fig. (d) are the best L-Shape fitting results from Alg.~\ref{Args:shape_fitting} with the results of Fig. (a) and Fig. (b). (best viewed in color.)}
	\label{figs:fitting1}
\end{figure}

\begin{figure}
	\centering
	\subfloat{\includegraphics[width= 3.6in]{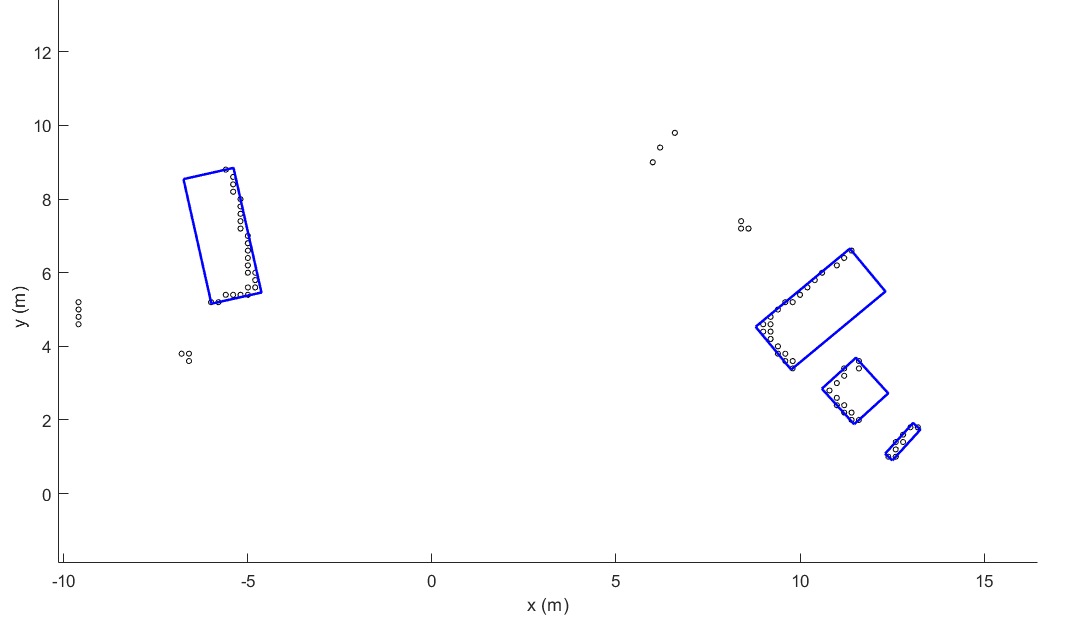}}
	\captionsetup{font={small}}
	\caption{The L-Shape fitting results for laser scan data and vehicle pose estimation. The blue boxes represent the L-Shape fitting result and also pose estimation of vehicles.}
	\label{figs:fitting2}
\end{figure}

Before performing L-Shape fitting, the laser scan data needs to be partitioned into different clusters. Fig~\ref{figs:DBSCAN_result} shows the segmentation result of 1 single scan data with the DBSCAN algorithm.
After the laser scan data been segmented into clusters, we use the fitting algorithms to search for the optimal L-Shape to fit the data points. Two different clusters which represent two different vehicles in two separate orientations are shown in Fig.~\ref{figs:fitting1}. Through the Alg.~\ref{Args:Vertexes searching} and Alg.~\ref{Args:corner_point_seeking} the key points are presented in Fig.~\ref{figs:fitting1}(a) and (b). The blue stars stand for the possible vertexes for L-Shape, and the red diamonds are the best corner points for each L-Shape in the circumstance of blue stars as vertexes of the L-Shape. With these key points' information and other results obtained from Alg.~\ref{Args:Vertexes searching} and Alg.~\ref{Args:corner_point_seeking} the optimal L-Shape for each cluster can be acquired from Alg.~\ref{Args:shape_fitting}. In Fig.~\ref{figs:fitting1}(c) and (d) the blue rectangle presents the optimal fitted L-Shape. \iffalse and also stand for the orientation of vehicles. \fi Fig.~\ref{figs:fitting2} shows the L-Shape fitting results of 1 single laser scan data and vehicle pose estimation. Each blue box is the best-fitted L-Shape obtained by fitting algorithms corresponding to each vehicle, and the directions of these rectangles are the orientation of vehicles.

It should be noted that the small clusters, with less than four points, are ignored in the implementation. Since these clusters are impossible to correspond to vehicles.
\subsection{Efficiency Evaluation}
The efficiency of the algorithm is evaluated by the computational time. There are approximately 3000 laser scans in the data set collected by the tested vehicle. Each laser range scan points is segmented into clusters and fitting algorithms are carried out on each cluster. The computational time is presented in Table~\ref{table:efficiency}. The calculations are implemented in MATLAB and run on a Windows laptop equipped with an Intel Core i5 CPU. The computational performance of the algorithm could be much better if it is implemented with a more efficient programming language such as C/C++ or on a more powerful platform.
\begin{table}
	\centering
	\caption{Computation Time of L-Shape Fitting}
	\label{table:efficiency}
	\begin{tabular}{ccc}
		\toprule
		Method		&Average (ms)   &Standard Deviation (ms)\\
		\midrule
		Our approach	&6.20		   &0.20\\
        CMU's method~\cite{CMU_LShape}~\footnote{Due to the difference of testing platform, the computation time is different from ~\cite{CMU_LShape}.}	&6.04	 &0.23\\
		\bottomrule	
	\end{tabular}       
\end{table}	
\section{Conclusion}
In this paper, we proposed a search-based L-Shape fitting approach. The algorithm can efficiently detect the optimal L-Shape fitting with 2D LiDAR data by finding the three key points of an L-Shape, that is two vertexes and one corner. The proposed approach does not need the scan's ordering/sequential information, therefore it allows fusions of raw laser data from multiple laser scanners.
Furthermore, this approach is capable of accommodating to
various criteria, which means the approach is not only suitable
for different fitting demands but also extensible for future applications. The experimental results show the correctness and efficiency of our algorithm.
\bibliographystyle{ieeetr}
\bibliography{Lshape}

\end{document}